\documentclass[10pt, a4paper]{article}

\usepackage[final]{lrec2026} 
\usepackage{amsmath}
\usepackage{booktabs}

\title{Quality and Agreement in Multilabel Emotion Annotation:\\A Case Study and Evaluation Framework}

\name{Emily Ohman \& Anna Koufakou} 

\address{Waseda University, Florida Gulf Coast University \\
         Tokyo, Japan; Fort Myers FL, USA \\
         ohman@waseda.jp, akoufakou@fgcu.edu\\
        }

\abstract{Emotion annotation is inherently subjective, yet most NLP pipelines still assume ``gold'' labels, typically produced by majority voting, and treat annotator variation as noise. In this paper, we present a multilabel emotion annotation case study and use it to examine how annotator behavior and aggregation choices affect both agreement estimates and downstream emotion classifiers. Rather than collapsing disagreement into a single label, we represent targets as soft vote-share labels (including an intensity-weighted variant) and evaluate models using both thresholded metrics (macro-/micro-F1) and probabilistic alignment (Bernoulli cross-entropy SoftBCE), alongside data-derived disagreement diagnostics. Across annotation regimes, we show that disagreement is structured and leaves measurable traces in model behavior: hard labels may maximize F1 metrics, while soft supervision yields predictions that better reflect empirical annotator variance and uncertainty. Our results provide practical guidance for designing, aggregating, and evaluating multilabel emotion datasets when multiple interpretations are plausible.
\\ \newline \Keywords{emotion annotations, annotator disagreements, perspectives, evaluation metrics} }

\begin{document}

\maketitleabstract

\section{Introduction}

The difficulty of annotating for emotions is widely acknowledged in the field of emotion detection and sentiment analysis \cite[see e.g.][]{andreevskaia2007clac,bermingham2009study,mohammad2016practical,emosentdiff,strapparava2010annotating,wiebe2005creating,ohman2021language}. 
Recent reviews also summarize the research landscape, available resources, challenges and gaps in this area \cite{plaza-del-arco-etal-2024-emotion,koufakou2025review}. In the following, we focus on specific aspects and challenges related to our work.

Overall, NLP research typically uses basic emotions \cite[see e.g.][]{ekman1971universals,plutchik1984emotions} or simplifies fine-grained frameworks \citep[see e.g.][]{cowen2017self,ohman2020emotion}, relying on classification models with a small set of largely disjoint labels. This facilitates model training and improves performance: for example, the GoEmotions \cite{demszky2020goemotions} experiments showed higher performance when the authors merged and remapped fine-grained categories to basic emotions (Ekman). These techniques, however, fail to capture the full, fine-grained, and often overlapping nature of human emotional experience.
In addition to discrete-category theories, affect is also commonly modeled in dimensional terms such as valence–arousal–dominance (VAD) and related circumplex accounts \cite{russell1977evidence}.

Equally notable is that much of the related research in NLP relies on the notion of ``gold'' labels, typically derived through majority voting among annotators. Due to the highly subjective nature of interpreting emotions, let alone emotions in a medium like text, humans usually do not agree on emotion labels. When annotators disagree, the corresponding records are often discarded. As \citet{plank-2022-problem} notes, ``human variation in labeling is often considered noise.'' Yet this raises an important question: \textit{is such variation truly noise}? Aside from cases of genuinely careless or inconsistent annotators, whose contributions may indeed warrant removal, disagreement among humans often indicates that an item is open to interpretation. In such cases, rather than enforcing a single ``correct'' label, we might instead expect NLP models to reflect this nuance in their predictions.

Agreement and alignment among annotators is measured by Inter-Annotator Agreement (IAA). There are several metrics used to assess IAA; most common are Cohen’s $\kappa$, Fleiss’ $\kappa$, and Krippendorff’s $\alpha$. The choice of IAA metric is influenced by various factors, for example the complexity of the annotation scheme (e.g. binary vs multi-label), the number of annotators, and how the work is assigned to the annotators. Regardless of the IAA metric, highly subjective tasks such as perceived emotions might lead to moderate or even low IAA. On the other hand, high agreement does not always mean high-quality annotations, for example, annotators could consistently agree on the wrong label. Besides the agreement between annotators, there might also be questions as to removing or filtering certain labels based on annotator effort or quality, which are not as straightforward to answer as we show in our case study (see Section \ref{case-study}).

Instead of eliminating disagreement, embracing it can reveal the range of emotional interpretations a text may evoke. \textit{Perspectivism} in NLP \cite{basile2021we,cabitza2023toward,frenda2025perspectivist} challenges the assumption of a single, objective ground truth and recognizes that multiple valid interpretations arise from annotators’ diverse backgrounds, experiences, and perspectives. By preserving disagreement, \textit{perspectivism} enables more inclusive and representative emotion recognition systems and reframes annotation not merely as labeling but as a means of exploring the full spectrum of emotions relevant to a task. In the past, only a few datasets released full individual annotator labels, e.g. \cite{demszky2020goemotions}, but this is starting to change\footnote{We release all processing code and derived annotation tables keyed by sentence index (without redistributing copyrighted text). The experiments can be reproduced by applying the scripts to a locally obtained copy of the novel. The code is available at \url{https://github.com/esohman/SoftBCE_Emotions}.}\cite{barz2025understanding,muhammad-etal-2025-brighter}. Additionally, there have been efforts to gain insights into the annotator disagreements, for example in environmental communications \cite{barz2025understanding}, while \citet{webergenzel2024varierrnliseparatingannotation} showed the challenges of distinguishing between human label variation that may be important to preserve and annotation errors.

To preserve human disagreement, rather than focusing on traditional evaluation metrics defined over gold or \textit{hard} labels, prior work has proposed the use of \textit{soft} labels, for example, see earlier SemEval Tasks \textit{Learning with Disagreement} \cite{leonardelli-etal-2023-semeval,leonardelli-etal-2025-lewidi} based mostly on data with binary or Likert scale labels. To the best of our knowledge, soft-label approaches have not yet been explored specifically for emotion annotation. 

In this paper, we present a multilabel emotion annotation case study and examine how soft labels can be used to preserve annotator disagreement, inform annotation quality assessment, and support appropriate NLP model evaluation. The case-study anchors the following contributions we make:
\begin{itemize}
    \item A critical review of annotator agreement metrics and common practice of assigning hard labels for NLP
    \item A multilabel emotion annotation case study illustrating real disagreement patterns and questions of annotator quality
    \item A soft-label framework that preserves annotator disagreement, together with an evaluation framework and metrics specifically aligned with soft labels in a multilabel setting, covering both emotion categories and intensity annotations.
\end{itemize}

Section~\ref{case-study} introduces the case study, followed by our methodology for aggregating multi-annotator labels in Section \ref{sec:targets_disagreement}, and then the evaluation framework for agreement analysis and downstream NLP experiments in Section \ref{sec:eval-pipeline}. Finally, we provide our concluding remarks in Section \ref{conclusions}.

\section{Case Study} \label{case-study}

We present a case study of three annotators annotating \citeauthor{hemingway1995old}'s \textit{The Old Man and the Sea} (in English) using an expanded version of Plutchik's wheel of emotions.

\begin{table}[!htbp]
\centering
\footnotesize
\setlength{\tabcolsep}{4pt}
\begin{tabularx}{\columnwidth}{lX}
\toprule
\textbf{Field} & \textbf{Value} \\
\midrule
Resource name & The Old Man and the Sea emotion annotations (TOMATS)\\
Resource type & Sentence-level multilabel emotion annotations with intensity \\
Language & English \\
Unit of annotation & Roughly sentence-level text segments \\
Size & 1{,}677 data points (sentences) \\
Annotators & 3 expert annotators \\
Label set (core) & \{anger, anticipation, disgust, fear, joy, sadness, surprise, trust\} \\
Additional labels & Optional free-text ``other'' emotions/notes (not modeled) \\
Intensity & Pre-selected-emotion rating on 0--10; rescaled to [0,1] for modeling \\
Annotation format & Spreadsheet (one row per item; up to 3 emotions + intensities + other) \\
\bottomrule
\end{tabularx}
\caption{Resource card for the TOMATS emotion annotation case study.}
\label{tab:resource_card}
\end{table}

\subsection{Dataset and Annotation Protocol}

All three annotators were research assistants of the first author and native speakers of English, majoring in NLP or adjacent fields. Annotators 2 and 3 received most of their education in the US, while Annotator 1 attended international schools with English as the primary language of instruction. Although demographically similar, women in their mid-20s, they differ in cultural upbringing and educational exposure, which we expected to influence annotation style.

Ideally, a broader demographic profile (for example, including older and male annotators) would have improved diversity, but the present study uses the available annotators to highlight how even superficially “similar” annotators can diverge substantially in emotion annotation.

The annotators were asked to first read the text in full before considering the emotions and then start from the beginning to label each sentence\footnote{roughly sentence level with minor splitting/merging where needed for coherence}  in context. They could assign multiple emotions or select ``no emotion.'' For the core annotation, annotators selected from the eight Plutchik emotions: \textit{anger, anticipation, disgust, fear, joy, sadness, surprise, trust}. Annotators could additionally enter free-text ``other'' emotions/notes when none of the core labels captured their interpretation. These ``other'' emotions are listed in Appendix \ref{app:b}.

We also introduced Plutchik’s dyad system to encourage reflection on secondary and tertiary emotions and allowed annotators to add additional categories or notes when they felt the taxonomy was insufficient. Annotators also rated the intensity of each selected emotion on a 0–10 scale. We used a 0–10 scale to provide annotators a familiar, high-resolution ordinal range that supports within-item comparisons (e.g., weak vs. strong evidence for a selected emotion). For modeling, we linearly rescaled intensities to [0,1] so they can be used directly as weights in vote-share aggregation while keeping all soft targets within the Bernoulli objective’s natural range.

\subsection{Annotator Behavior and Disagreement}

\begin{table}[!htbp]
\centering
\begin{tabular}{llll}
\hline
\textbf{Sum} & \textbf{primary} & \textbf{secondary} & \textbf{tertiary} \\
\hline
\textbf{A1}           & 1593    & 1258      & 443      \\
\textbf{A2}           & 184     & 13        & 3        \\
\textbf{A3}           & 1369    & 1200      & 566     \\
\hline
\end{tabular}
\caption{Annotation counts for each annotator}
\label{annocount}
\end{table}

Table \ref{annocount} shows the number of data points annotated by each annotator. We can see that Annotator 1 and Annotator 3 annotated emotions at similar rates across the 1677 sentences and assigned comparable numbers of secondary and tertiary labels. Their primary-emotion agreement is 42\%, but agreement rises to 62\% when considering any overlapping emotion label. 

Annotator 2 presents a different case. They only considered roughly 10\% of the data to have any emotion-association and almost no secondary or tertiary emotion associations (13 and 3 respectively). A closer look at their annotations further reveal that not only did they not annotate many data points at all, the data points they did annotate were not, contrary to expectations, data points with clear high-intensity emotion-linked content. Taken together with similarly sparse completion patterns observed in other research tasks, this points to limited protocol engagement rather than an alternative emotional reading.

\begin{table}[!htbp]
\centering
\begin{tabular}{l r}
\hline
\textbf{Emotion} & \textbf{Krippendorff's $\alpha$} \\
\hline
Anger        & 0.1674 \\
Anticipation & 0.2160 \\
Disgust      & 0.1632 \\
Fear         & 0.1599 \\
Joy          & 0.1984 \\
Sadness      & 0.2793 \\
Surprise     & 0.2075 \\
Trust        & 0.0920 \\
\hline
\end{tabular}
\caption{Krippendorff's $\alpha$ for binary presence/absence of each Plutchik emotion.}
\label{tab:kripp}
\end{table}

Preliminary agreement scores calculated using Krippendorff's $\alpha$ in Table \ref{tab:kripp} indicate low agreement. We include these metrics to highlight how unsuitable they are for multilabel emotions despite being commonly requested by reviewers.

\begin{figure}
    \centering
    \includegraphics[trim={0.4cm 0.5cm 0.9cm 1cm}, clip, width=0.96\linewidth]{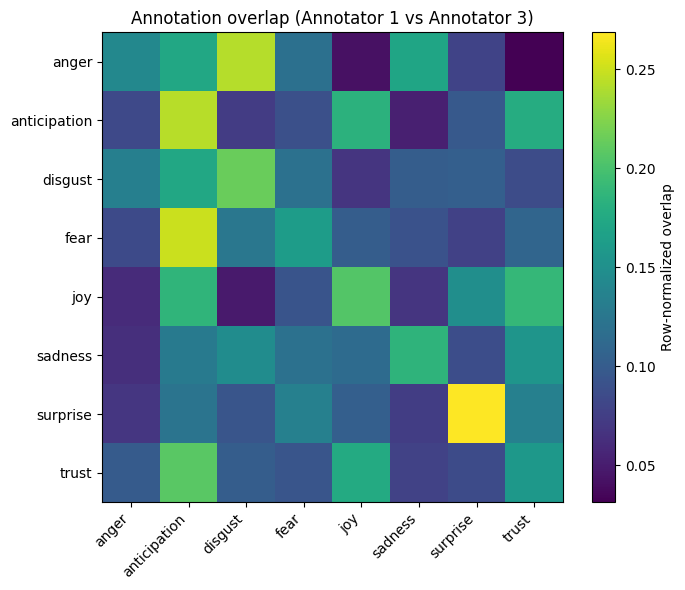}
    \caption{Annotation overlap (A1 vs A3)}
    \label{fig:anno_conf}
\end{figure}
To contextualize model confusion patterns, we additionally compute an annotation overlap matrix that captures systematic disagreement between annotators. We construct an emotion–emotion co-occurrence matrix (Fig. \ref{fig:anno_conf}) by counting how often an emotion selected by one annotator overlaps with an emotion selected by another annotator for the same item, allowing up to three labels per annotator. The resulting matrix is row-normalized and visualized as a heatmap. We can see that \textit{fear} is often confused with \textit{anticipation}, and \textit{anger} with \textit{disgust} something that was also noted by e.g. \citet{ohman2024emotionarcs}.
This representation does not encode correctness, but rather reflects structured ambiguity in the annotation process itself. 
Appendix \ref{app:a} shows intra-annotator emotion overlap matrices.

\subsection{Agreement Analysis and Implications}

    To assess annotation reliability, we compute pairwise agreement scores between the three annotators across all eight Plutchik emotions. Because the task is multilabel, each instance may exhibit zero or more emotions; classical inter-annotator agreement coefficients (e.g., Cohen’s~$\kappa$ or Fleiss’~$\kappa$) are therefore not directly applicable, as they assume mutually exclusive categorical judgments. Instead, we adopt a per-emotion binary evaluation framework commonly used in multilabel affect annotation, where agreement is computed independently for each emotion (e.g., \citealp{doi:10.1111/j.1467-8640.2012.00460.x}; \citealp{mohammad2017emotion}) (see also \citealp{artstein2008inter} for a general discussion of agreement in non-exclusive annotation tasks).

For each annotator pair $(a,b)$, for emotion $e$, and item $i$, we consider binary labels
\begin{equation}
  y^{(a)}_{i,e},\; y^{(b)}_{i,e} \in \{0,1\},
\end{equation}
indicating the presence or absence of emotion $e$. We compute the per-emotion F1 score as this metric directly captures how consistently annotators agree on the presence of an emotion and is less sensitive to extreme class imbalance than accuracy.
From the set of eight per-emotion F1 scores, we report:
\begin{itemize}
    \item \textbf{macro-F1}: the unweighted mean of the per-emotion F1 scores;
    \item \textbf{micro-F1}: the F1 score computed after flattening all $8\times N$ binary decisions into a single contingency table;
    \item \textbf{Per-emotion~F1}: the individual agreement scores for each Plutchik emotion.
\end{itemize}

Annotator~2 left the majority of items unannotated (``none''). Here, ``none''/blank entries denote no provided label (nonresponse/placeholder) rather than an explicit `no emotion' judgment, which was available as a selectable option. Treating such blanks as negative judgments would artificially deflate agreement. To distinguish disagreement from missing information, we therefore measure agreement under two conditions: (i) on the full dataset, where blank entries count as negatives, and (ii) on a filtered subset consisting only of items where Annotator~2 produced at least one genuine emotion label. In the latter setting, we treat all other entries for Annotator~2 as missing.

\begin{table*}[t]
\centering
\small
\begin{tabular}{l l r r r r}
\hline
Pair & Setting & macro-F1 & micro-F1 & Avg. Per-emotion F1 & Min/Max Per-Emotion F1 \\
\hline
A1--A2 & Full     & 0.0788 & 0.0882 & 0.0795 & 0.0093 / 0.1760 \\
A1--A2 & Filtered & 0.3949 & 0.4857 & 0.3932 & 0.1053 / 0.7015 \\
A1--A3 & Full     & 0.3794 & 0.3952 & 0.3804 & 0.2857 / 0.4980 \\
A1--A3 & Filtered & 0.4899 & 0.5187 & 0.4870 & 0.3051 / 0.6410 \\
A2--A3 & Full     & 0.0829 & 0.0899 & 0.0820 & 0.0058 / 0.1469 \\
A2--A3 & Filtered & 0.3797 & 0.4664 & 0.3830 & 0.0455 / 0.7706 \\
\hline
\end{tabular}
\caption{Summary comparison of pairwise annotator agreement in the full vs.~filtered subsets.}
\label{tab:comparison}
\end{table*}

In Table \ref{tab:comparison}, we observe substantial differences in pairwise annotator agreement depending on whether the analysis is performed on the full dataset or restricted to the subset of items that Annotator~2 actually annotated with a genuine emotion label. On the full dataset, agreement between Annotator~2 and the other annotators is extremely low (macro-F1 $\approx$ 0.08), reflecting the fact that Annotator~2 left most items unannotated and thus systematically disagreed with the others. However, when we restrict the evaluation to the 184 items that Annotator~2 labelled, agreement rises sharply (macro-F1 $\approx$ 0.38-0.40), approaching the level of agreement observed between the two reliable annotators (Annotator~1 and Annotator~3: macro-F1 $\approx$ 0.49). This indicates that Annotator~2 is not inherently inconsistent; rather, the majority of their missing labels should be treated as missing data rather than as true negatives. Consequently, for all downstream modeling we treat Annotator~2’s missing annotations as missing information rather than evidence of emotion absence, and only incorporate their annotations on items they explicitly rated.

As a sidenote, it should not be ignored that the low agreement scores are in part because the annotators, Annotator~3 in particular, utilized the "other emotion" column. For example, in one case Annotator~1 annotated the primary emotion as \textit{anticipation} and \textit{fear} as the secondary emotion. For that same data point Annotator~3 had left the primary slot empty and added \textit{pessimism} as the other emotion. It could be argued that \textit{pessimism} is indeed an overlap of \textit{anticipation} and \textit{fear}. 

In another example, the sentence \textit{``Chew it well, he thought, and get all the juices."} occurs during Santiago’s (the protagonist) prolonged struggle at sea, he eats a small fish raw to maintain his strength. The internal monologue reflects pragmatic self-discipline and physical endurance rather than pleasure, emphasizing survival under extreme deprivation. Annotator~1 had selected \textit{disgust, anger, anticipation} (in that order) and Annotator~3 selected \textit{anger, anticipation}, and added \textit{determination}. Again, the annotations seem aligned with a human interpretation, but harder to accurately show quantitatively.

\section{Aggregating multi-annotator labels}
Here, we describe our methodology for constructing hard and soft targets from human multilabel annotations. 

Each instance is annotated by up to three annotators, who may select multiple emotions and optionally provide intensity ratings. Because Annotator~2 labels a smaller subset of instances, we treat missing annotations as missing rather than as explicit negative votes. For each instance $i$ and emotion $k$, we compute a soft target as the empirical vote share among contributing annotators:
\begin{equation}
\tilde{y}_{i,k}=\frac{1}{|\mathcal{A}_i|}\sum_{a\in\mathcal{A}_i}\mathbf{1}[k\in S_{a,i}]
\end{equation}
where $S_{a,i}$ is annotator $a$'s selected label set and $\mathcal{A}_i$ denotes annotators who provided at least one label for instance $i$. This produces multilabel targets $\tilde{y}_{i,k}\in[0,1]$ that can be interpreted as per-label Bernoulli probabilities.

To incorporate intensity information, we additionally define an intensity-weighted soft target. For each annotator, selected emotions are weighted by their reported intensity (scaled to $[0,1]$), and aggregated across contributing annotators. This preserves graded signal about strength of affect while still producing multilabel targets in $[0,1]$ suitable for Bernoulli objectives.

\subsection{Targets and disagreement diagnostics}
\label{sec:targets_disagreement}

Let $y^{(a)}_{i,k}\in\{0,1\}$ denote whether annotator $a$ assigned emotion $k$ to instance $i$, and let $\mathcal{A}_i$ denote the set of annotators who provided at least one label for instance $i$ (i.e., missing annotations are treated as missing). We define the per-label vote share
\begin{equation}
p_{i,k}=\frac{1}{|\mathcal{A}_i|}\sum_{a\in\mathcal{A}_i} y^{(a)}_{i,k}
\label{eq:vote_share}
\end{equation}
which yields multilabel soft targets $p_{i,k}\in[0,1]$ that can be interpreted as Bernoulli probabilities. Hard targets are obtained by union-of-labels over contributing annotators:
\begin{equation}
y^{\mathrm{hard}}_{i,k}=\mathbf{1}\!\left[p_{i,k}>0\right]
\label{eq:hard_targets}
\end{equation}

To characterize ambiguity in the annotations independently of any model, we compute two data-derived disagreement measures from annotator label sets. First, we use mean per-label Bernoulli variance,
\begin{equation}
D_{\mathrm{var}}(i)=\frac{1}{K}\sum_{k=1}^{K} p_{i,k}\big(1-p_{i,k}\big)
\label{eq:dvar}
\end{equation}
which is maximized when annotators split evenly on a label and minimized when they agree ($p_{i,k}\in\{0,1\}$). Second, we compute mean pairwise Jaccard disagreement between annotators' label sets,
\begin{equation}
D_{\mathrm{Jac}}(i)
=1-\frac{1}{\binom{|\mathcal{A}_i|}{2}}
\sum_{a<b}\frac{|S_{a,i}\cap S_{b,i}|}{|S_{a,i}\cup S_{b,i}|}
\label{eq:djac}
\end{equation}
where the sum is taken over all annotator pairs $(a,b)$ contributing to instance $i$. These measures are computed from annotations alone and are therefore constant across model variants trained under the same annotator-inclusion regime.

Because our downstream models operate over the 8 Plutchik emotions, we filter free-text ``other'' labels and compute soft-label distributions over this fixed set. This effectively reallocates probability mass from custom labels to the remaining categories via renormalization, which can reduce apparent disagreement (lower Dvar) in cases where annotators expressed qualitatively different states outside the taxonomy. Future work could mitigate this by introducing an explicit ``other'' class, mapping free-text labels into a hierarchical taxonomy, or analyzing Dvar both with and without the “Other” bucket.

\subsection{Bernoulli cross-entropy for multilabel soft targets}
\label{sec:softbce}

Because our task is multilabel, we model each emotion as an independent Bernoulli variable. Let $\hat{p}_{i,k}=\sigma(z_{i,k})$ denote the model's predicted probability for label $k$ (with logit $z_{i,k}$). We quantify probabilistic alignment between model predictions and the aggregated soft targets $p_{i,k}$ using Bernoulli cross-entropy (binary cross-entropy; BCE):
\begin{multline}
\mathrm{SoftBCE}(i,k)
= -\Big(p_{i,k}\log \hat{p}_{i,k} + \\
\big(1-p_{i,k}\big)\log\big(1-\hat{p}_{i,k}\big)\Big)
\end{multline}
\label{eq:softbce}
We report SoftBCE averaged across labels and instances as an auxiliary model--annotation alignment metric; lower values indicate closer alignment. In addition to SoftBCE, downstream models are evaluated using macro-/micro-F1 with decision thresholds tuned on a validation split. Similar cross-entropy-style diagnostics have been used in prior disagreement-oriented work (e.g., \citealp{leonardelli-etal-2023-semeval, leonardelli-etal-2025-lewidi}), but our setting differs in that targets are multilabel Bernoulli vote shares (with missingness treated explicitly), rather than single-label distributions.

Concrete stratified examples of the resulting hard and soft targets (low/medium/high disagreement) are provided in Table~\ref{tab:soft_hard_examples_stratified}.

\section{Evaluation Pipeline}\label{sec:eval-pipeline}

Our evaluation pipeline is designed to examine how annotation behavior and disagreement structure affect both agreement estimates and downstream emotion classification. Rather than optimizing predictive performance, the goal is to assess how different assumptions about annotation completeness and disagreement propagate through the modeling process.

The pipeline consists of four stages.

\paragraph{(1) Label extraction and harmonization.}
Annotations are provided in separate spreadsheet sheets, one per annotator, with up to three Plutchik emotion labels per item and optional free-text notes. For each annotator $a$ and item $i$, we extract the set of explicitly assigned Plutchik emotions, discarding ``none'', blanks, and non-Plutchik entries. This yields a mapping $(i) \rightarrow \{ e_1, e_2, \ldots \}$, that reflects only positive emotion assignments.

\paragraph{(2) Multilabel representation.}
We convert the extracted labels into a binary multilabel tensor
\[
Y \in \{0,1\}^{A \times N \times K}
\]

where $Y_{a,i,k}=1$ indicates that annotator $a$ assigned emotion $k$ to item $i$. This representation treats each annotator as an independent multilabel classifier and supports both pairwise agreement analysis and model training. Missing annotations (i.e., annotator--item pairs where no labels were provided) are encoded as empty label sets and tracked via a contribution mask, which is used in the next stage when aggregating vote shares over the set of contributing annotators $\mathcal{A}_i$.

\paragraph{(3) Agreement analysis with missing-label filtering.}
Because one annotator exhibits systematic sparsity, interpreting blank entries as negative judgments would conflate missingness with disagreement. We therefore compute agreement under two conditions:
\begin{enumerate}
    \item \textbf{Full dataset:} all missing labels are treated as ``no emotion''.
    \item \textbf{Filtered dataset:} items for which Annotator~2 assigned at least one genuine Plutchik emotion are retained; all other items are removed for all annotators.
\end{enumerate}
This comparison allows us to disentangle structural disagreement caused by missing labels from genuine divergence in emotional interpretation.

\paragraph{(4) Downstream modeling and evaluation.}
To assess whether differences in annotation behavior propagate into downstream models, we evaluate a small set of representative classifiers under different annotation regimes. The goal is not to optimize predictive performance, but to use modeling behavior as a diagnostic lens on annotation consistency. We further introduce an evaluation framework for soft labels in multilabel emotion annotation and demonstrate its practical application.

We consider four classifier families:
\begin{enumerate}
  \renewcommand{\labelenumi}{(\alph{enumi})}
  \item a transformer-based multilabel classifier trained with Bernoulli cross-entropy (binary cross-entropy) on aggregated hard targets;
  \item the same architecture trained with Bernoulli cross-entropy using vote-share soft targets $\tilde{y}_{i,k}\in[0,1]$, thereby preserving annotator disagreement in the supervision signal;
  \item the same architecture trained with intensity-weighted soft targets, where vote shares are modulated by annotator-provided intensity scores (scaled to $[0,1]$);
  \item a one-vs-rest SVM baseline using frozen transformer embeddings.
\end{enumerate}
Each model is trained under two annotation conditions: using all available annotators, and using only the two annotators identified as reliable by the filtered agreement analysis. We report both thresholded classification metrics (macro-/micro-F1) and probabilistic alignment metrics. Data-derived disagreement measures (e.g., $D_{\mathrm{var}}$ and $D_{\mathrm{Jac}}$) are computed from annotations only, while SoftBCE denotes the Bernoulli cross-entropy between model probabilities and the soft targets and is used as an auxiliary model--annotation alignment metric (see subsections \ref{sec:targets_disagreement} and \ref{sec:softbce} respectively).

\subsection{Training objectives}

We train all transformer models with a multilabel Bernoulli objective (\texttt{BCEWithLogitsLoss}). In the hard-label condition, targets are binary vectors aggregated from annotator label sets. In the soft-label conditions, we replace binary targets with either (i) vote-share soft targets $\tilde{y}$ or (ii) intensity-weighted soft targets, yielding a soft-target variant of BCE that exposes annotator uncertainty and graded affect strength during training. 
Examples are stratified by $D_{\mathrm{Jac}}$ tertiles (computed from annotator label sets).

\begin{table*}[!htbp]
\centering
\small
\setlength{\tabcolsep}{6pt}
\renewcommand{\arraystretch}{1.2}
\begin{tabular}{p{0.3\linewidth} p{0.15\linewidth} p{0.2\linewidth} p{0.2\linewidth}}
\hline
Example text & Annotations & Hard labels & Soft labels \\
\hline

\multicolumn{4}{l}{\textit{Low disagreement}}\\
\hline
``That's very kind of you,'' the old man said.
&
A1: Sa, Su, Tr 
\newline
A2: Jo, Tr 
\newline
A3: Jo, Tr 
&
$[0,0,0,0,1,1,1,1]$
&
$[0,0,0,0,\tfrac{2}{3},\tfrac{1}{3},\tfrac{1}{3},1]$
\\
``The bird is a great help,'' the old man said.
&
A1: Jo, Tr 
\newline
A2: Jo \newline
A3: Jo, Su 
&
$[0,0,0,0,1,0,1,1]$
&
$[0,0,0,0,1,0,\tfrac{1}{3},\tfrac{1}{3}]$
\\
\hline

\multicolumn{4}{l}{\textit{Medium disagreement}}\\
\hline
He was gone and the old man felt nothing.
&
A1: Sa 
\newline
A2: Sa 
\newline
A3: Su 
&
$[0,0,0,0,0,1,1,0]$
&
$[0,0,0,0,0,\tfrac{2}{3},\tfrac{1}{3},0]$
\\
Come up easy and let me put the harpoon into you.
&
A1: Ant, Jo, Tr 
\newline
A2: Ant \newline
A3: Ant, Di, Jo 
&
$[0,1,1,0,1,0,0,1]$
&
$[0,1,\tfrac{1}{3},0,\tfrac{2}{3},0,0,\tfrac{1}{3}]$
\\
\hline

\multicolumn{4}{l}{\textit{High disagreement}}\\
\hline
``I think perhaps I can too. But I try not to borrow. First you borrow. Then you beg.''
&
A1: Fe, Sa, Tr 
\newline
A2: Di 
\newline
A3: Ant, Jo, Tr 
&
$[0,1,1,1,1,1,0,1]$
&
$[0,\tfrac{1}{3},\tfrac{1}{3},\tfrac{1}{3},\tfrac{1}{3},\tfrac{1}{3},0,\tfrac{2}{3}]$
\\
``They say his father was a fisherman. Maybe he was as poor as we are and would understand.''
&
A1: Ant, Jo, Sa 
\newline
A2: Tr 
\newline
A3: Su, Tr 
&
$[0,1,0,0,1,1,1,1]$
&
$[0,\tfrac{1}{3},0,0,\tfrac{1}{3},\tfrac{1}{3},\tfrac{1}{3},\tfrac{2}{3}]$
\\
\hline
\end{tabular}

\caption{
Stratified examples illustrating how hard and soft targets are derived from human annotations. Hard labels are the union-of-labels over annotators (1 if any annotator selected the emotion).
Soft labels are vote-share targets (fraction of contributing annotators selecting each emotion).
Vectors follow the fixed emotion order:
[Ang:anger, Ant:anticipation, Di:disgust, Fe:fear, Jo:joy, Sa:sadness, Su:surprise, Tr:trust].
}
\label{tab:soft_hard_examples_stratified}
\end{table*}
We report both thresholded classification metrics (macro-/micro-F1) and probabilistic alignment metrics. While F1 evaluates discrete decisions after thresholding, SoftBCE evaluates the full predicted probability vector against the aggregated soft targets; improvements in SoftBCE therefore indicate better-calibrated probabilities that more faithfully reflect inter-annotator variation.

Table \ref{tab:model_details} lists implementation details for reproducibility.

\begin{table}[!htbp]
\centering
\footnotesize
\setlength{\tabcolsep}{4pt}
\begin{tabularx}{\columnwidth}{lX}
\hline
\textbf{Component} & \textbf{Setting} \\
\hline
Backbone & \texttt{bert-base-uncased}  \\
Tokenizer & \texttt{AutoTokenizer} (uncased WordPiece) \\
Input length & Max length 256; truncation; padding to \texttt{max\_length} \\
Optimizer & AdamW, learning rate $2\times 10^{-5}$ \\
Batch size & 16 \\
Epochs & Up to 10 with early stopping \\
Early stopping & Patience 2 on validation macro-F1; keep best checkpoint (min $\Delta=10^{-4}$) \\
Threshold selection & Tune global threshold $t$ on validation macro-F1; grid $t \in \{0.10,0.15,\ldots,0.90\}$ \\
Data split & Random 80/20 train/validation split; shuffle=True \\
Random seed & 42 (\texttt{numpy} + \texttt{torch}) \\
SVM baseline & One-vs-rest LinearSVC on frozen BERT [CLS] embeddings (embedding batch size 32; other params default) \\
\hline
\end{tabularx}
\caption{Model and training configuration used in all transformer experiments.}
\label{tab:model_details}
\end{table}
All results are from a single run with fixed seed; we did not perform multi-seed averaging.

\begin{table*}[!htbp]
\centering
\small
\setlength{\tabcolsep}{7pt}
\renewcommand{\arraystretch}{1.15}
\begin{tabular}{llccccccc}
\hline
Setting & Model & Thr. & macro-F1 & micro-F1 & SoftBCE & $D_{\text{var}}$ & $D_{\text{Jac}}$ & $\overline{A}$ \\
\hline
With A2
& Hard labels             & 0.35 & 0.616 & 0.625 & 0.681 & 0.066 & 0.708 & 1.98 \\
With A2
& Soft labels (uniform)   & 0.10 & 0.589 & 0.599 & 0.577 & 0.066 & 0.708 & 1.98 \\
With A2
& Soft labels (intensity) & 0.10 & 0.607 & 0.617 & 0.548 & 0.066 & 0.708 & 1.98 \\
With A2
& SVM baseline & - & 0.503 & 0.519 & - & - & - & - \\
\hline
Without A2
& Hard labels             & 0.15 & 0.606 & 0.616 & 0.717 & 0.065 & 0.707 & 1.85 \\
Without A2
& Soft labels (uniform)   & 0.10 & 0.588 & 0.597 & 0.564 & 0.065 & 0.707 & 1.85 \\
Without A2
& Soft labels (intensity) & 0.10 & 0.591 & 0.600 & 0.578 & 0.065 & 0.707 & 1.85 \\
Without A2
& SVM baseline & - & 0.501 & 0.517 & - & - & - & - \\
\hline
\end{tabular}

\caption{
Downstream results for BERT under different supervision targets and annotator-inclusion regimes.
Thr.\ is the tuned decision threshold (validation set).
SoftBCE is the Bernoulli cross-entropy between predicted probabilities and aggregated soft targets (lower is better).
$D_{\text{var}}$ denotes mean per-label Bernoulli variance $\frac{1}{K}\sum_k \tilde{y}_{k}(1-\tilde{y}_{k})$,
$D_{\text{Jac}}$ denotes mean pairwise Jaccard disagreement (both data-derived; higher means more disagreement),
and $\overline{A}$ is the average number of contributing annotators per item.
}
\label{tab:mod_comp}
\end{table*}

\subsection{Evaluation}
Table~\ref{tab:mod_comp} compares BERT models trained with hard targets versus two soft-target variants, under settings that either include or exclude Annotator~2. Hard-label training yields the highest F1 overall when Annotator~2 is included (macro-F1 = 0.616; micro-F1 = 0.625), but the intensity-weighted soft-label model remains competitive (macro-F1 = 0.607; micro-F1 = 0.617) while achieving substantially lower SoftBCE (0.548 vs.\ 0.681), indicating markedly better alignment between predicted probabilities and the aggregated soft targets. Uniformly weighted soft labels reduce F1 more noticeably, though they still improve SoftBCE relative to hard training. Excluding Annotator~2 slightly decreases F1 across objectives and also increases SoftBCE for the hard-label model, suggesting that Annotator~2 contributes useful signal despite partial coverage. Finally, the disagreement measures $D_{\text{var}}$ and $D_{\text{Jac}}$ are stable within each annotation regime, as expected because they are computed from the annotations themselves rather than model outputs; they confirm that the underlying level of annotator divergence is comparable across the compared training objectives.

Because macro-/micro-F1 are computed after binarizing predictions, they depend on a tuned decision threshold and primarily reflect ranking performance around that cutoff. SoftBCE, in contrast, evaluates the full vector of predicted probabilities against the aggregated soft targets without thresholding. The two metrics therefore capture complementary aspects of model quality: thresholded classification performance versus probabilistic alignment with annotator uncertainty.

Incorporating intensity information improves the soft-label objective. Intensity-weighted soft labels yield higher macro- and micro-F1 than uniformly weighted soft labels in both the \textit{with A2} and \textit{without A2} settings, and they also produce markedly lower SoftBCE. This suggests that intensity ratings provide a useful inductive bias that sharpens the soft targets by distinguishing weak affective cues from strong ones, improving both downstream decisions and probabilistic alignment with annotator uncertainty. Overall, these results support the use of soft labels as a diagnostic tool for understanding how annotation regimes propagate into model behavior, while highlighting that additional annotation detail does not automatically translate into improved downstream performance.

\begin{figure}
    \centering
    \includegraphics[trim={0.4cm 0.5cm 0.9cm 1cm}, clip, width=0.95\linewidth]{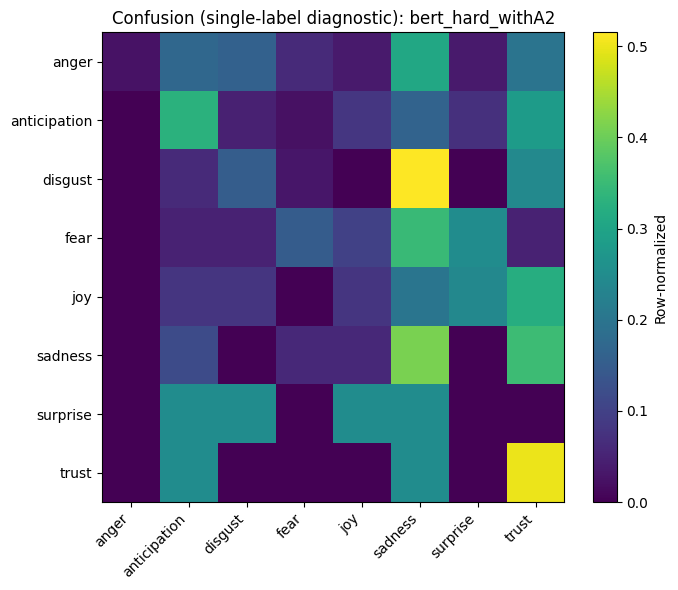}
    \caption{Confusion matrix for BERT with hard labels including A2 labels}
    \label{fig:confusion}
\end{figure}
\subsection{Confusion}
Figure~\ref{fig:confusion} presents a row-normalized single-label confusion matrix for the results from training BERT on hard labels with all annotators. Several emotions exhibit strong diagonal dominance, most notably \textit{sadness} and \textit{trust}, indicating that when these emotions are the primary label, the model predicts them consistently. The most prominent off-diagonal confusion occurs between \textit{disgust} and \textit{sadness}, with a substantial proportion of \textit{disgust}-labeled instances predicted as \textit{sadness}, suggesting overlap in negative-valence affective cues. This has also been shown in previous work \citep{rossi2025combining}. 
Other emotions, such as \textit{anger} and \textit{anticipation}, display more diffuse prediction patterns without a single dominant confusion partner. Overall, the matrix reveals clusters of related emotions rather than random error, supporting its use as a diagnostic tool for identifying structured affective overlap rather than discrete misclassification.

\subsection{Error Analysis}

To complement aggregate F1 scores, we inspect structured error patterns using a row-normalized single-label confusion matrix, obtained by projecting multilabel outputs to a dominant label per instance. Confusions are not random: the largest off-diagonal mass occurs between semantically and valence-adjacent emotions, including \emph{sadness} vs.\ \emph{fear}, \emph{trust} vs.\ \emph{anticipation}, and \emph{anger} vs.\ \emph{disgust}. These patterns align with the dataset's disagreement diagnostics ($D_{\mathrm{var}}$, $D_{\mathrm{Jac}}$), suggesting that model errors concentrate in regions of the label space that are also ambiguous for annotators. Predictive performance is weakest for rare emotions (e.g., \emph{surprise}), consistent with challenges in multilabel learning under extreme imbalance.

\section{Conclusions}\label{conclusions}

This paper examined how annotation design choices and annotator reliability affect downstream modeling for multilabel emotion classification. Rather than treating annotation as a static preprocessing step, we used it as an object of analysis and asked how different aggregation strategies and training objectives interact with empirically observed disagreement. Across experiments, annotation decisions produced consistent, measurable differences in both thresholded performance (macro-/micro-F1) and probabilistic model-target alignment (Bernoulli cross-entropy or SoftBCE).

The downstream results suggest that the benefits of soft supervision depend on what is being optimized. With all annotators included, hard-label training achieved the highest macro- and micro-F1 in our setup. Soft-label objectives, however, produced markedly lower SoftBCE while remaining broadly competitive in F1. In particular, intensity-weighted soft targets reduced SoftBCE the most under the full-annotator regime and narrowed the F1 gap to hard labels, indicating closer alignment with the uncertainty encoded in the aggregated targets. When we excluded Annotator~2, both hard and soft variants showed slightly lower F1, while the disagreement diagnostics ($D_{\mathrm{var}}$, $D_{\mathrm{Jac}}$) remained essentially unchanged, as expected for measures computed from the annotations alone.

These findings clarify a common claim in disagreement-aware learning. Preserving disagreement does not necessarily translate into higher F1, but it can yield probability estimates that better reflect the empirical label uncertainty. For affective tasks, where multiple interpretations can be simultaneously plausible and downstream applications often consume probabilities rather than hard decisions, this distinction is practically important. In this sense, SoftBCE provides a useful complementary lens to standard F1 when evaluating uncertainty-aware training signals.

At the same time, our results highlight limits to how additional annotation detail translates into downstream gains. Intensity ratings are not automatically helpful: their impact depends on how consistently annotators use the scale and on how intensities are normalized and aggregated. In this case study, incorporating intensity improved model-target alignment substantially but did not consistently dominate thresholded metrics, suggesting that intensity can encode graded uncertainty even when it does not directly improve binary decisions. This points to straightforward extensions, including annotator-specific calibration, learned normalization, and objectives that explicitly model individual differences in scale use.

The main contribution of this paper is methodological. Using downstream models as a diagnostic tool, together with disagreement measures computed directly from the labels, provides a practical way to surface design trade-offs in emotion annotation and to distinguish ambiguity in the data from annotator-specific behavior. Applying the same pipeline to additional texts and domains would help establish when these patterns generalize and when they are corpus- or task-specific. 

We anticipate that applying this pipeline to short-form or informal text (e.g., posts or comments) will increase both overlap among emotions and annotator disagreement due to reduced context and higher ambiguity. Our approach is designed for precisely these conditions: missingness-aware aggregation separates nonresponse from negative judgments, and SoftBCE measures distributional alignment when “gold” labels are intrinsically uncertain. Future work can further test robustness under larger, crowd-sourced annotator pools and examine how disagreement patterns vary with domain and annotator diversity.

In conclusion, the results support treating annotation and modeling as coupled choices rather than sequential steps. By making disagreement explicit in both target construction and evaluation, and by reporting both thresholded and probabilistic alignment metrics, we provide a transparent framework for analyzing and improving multilabel emotion annotation practices.

\section*{Limitations}

This work is intentionally framed as a case study, and several limitations are worth noting. First, the annotation effort involves a small number of annotators and a single literary text. This setting enables detailed qualitative and quantitative analysis of annotator behavior, but it limits the generalizability of absolute agreement values and downstream performance. Our conclusions therefore emphasize \emph{patterns} of annotation behavior and their modeling consequences rather than population-level estimates of annotator reliability.

Second, the annotators were not drawn from a fully representative population. Although all were native English speakers, they were relatively homogeneous and affiliated with the same research group. This reduces demographic variance, but also makes the observed divergence in annotation density and engagement noteworthy: substantial differences in labeling behavior emerged even within a narrow pool. Future work should test whether similar patterns hold with larger and more heterogeneous annotator groups. 

Third, one annotator exhibited systematically sparse labeling behavior, which complicates both agreement measurement and target construction. While we address this through missingness-aware aggregation (treating non-annotations as missing rather than negative votes) and filtered analyses, the degree of missingness underscores the need for clearer guidelines and more explicit handling of abstentions in emotion annotation workflows. In particular, low agreement scores can reflect missing labels rather than conceptual disagreement, but distinguishing these cases is not always straightforward in practice.

Fourth, while we use Bernoulli cross-entropy (SoftBCE) as a model-target alignment metric for multilabel soft targets, there is no single universally accepted evaluation measure for soft labels especially for multilabel settings, and different metrics emphasize different aspects of disagreement. Prior work has pointed out limitations of cross-entropy-style metrics in some soft-label settings \citep{rizzi2024soft}, and recent shared-task work has also explored alternatives such as Manhattan and Wasserstein distances \citep{leonardelli-etal-2025-lewidi}. We therefore treat SoftBCE as an auxiliary diagnostic and report complementary data-derived disagreement measures; evaluating additional soft-label metrics for multilabel emotion annotation remains an important direction for future work.

Fifth, although the annotation scheme includes additional (rare) emotion categories beyond the main set analyzed here, we did not model these extra labels. Incorporating very low-frequency categories without destabilizing training or evaluation is non-trivial, and developing principled strategies for handling rare emotions (e.g., hierarchical labels, grouping, or targeted re-annotation) is a clear next step. Similarly, the intensity score reliability and coherence could be improved by adapting a best-worst-scaling approach \citep[as demonstrated by][]{kiritchenko-mohammad-2017-best}.

Finally, the classification models evaluated in this paper are deliberately simple and are used primarily as analytical tools rather than optimized systems. The goal is not to establish state-of-the-art performance, but to examine how different annotation regimes affect downstream modeling and evaluation. More sophisticated architectures or training strategies may yield higher performance, but they would not change the core methodological point: annotation behavior and aggregation choices can measurably shape both model outputs and how those outputs should be evaluated.

\section*{Ethical Considerations}

This paper is primarily methodological: we analyze a small multilabel emotion annotation case study to understand how aggregation and disagreement-aware modeling choices affect evaluation and interpretation. The work does not involve user-generated content, sensitive personal data, or human subjects beyond a limited internal annotation exercise. Nonetheless, automated emotion recognition has broader societal implications that warrant explicit discussion.

Emotions are not directly observable from text, and the same utterance can support multiple plausible readings depending on context, culture, and reader perspective. Text-only emotion models are therefore best understood as predicting \emph{annotated interpretations} rather than latent ``true'' emotions. This is especially important in settings where other signals (e.g., conversational context, prosody, facial expression, or situational metadata) are unavailable. Our results reinforce this point: disagreement is structured and often reflects genuine ambiguity rather than annotator error. Treating majority-voted labels as ``gold'' can mask uncertainty and overstate model reliability.

Emotion recognition models can be misapplied to surveillance, profiling, workplace monitoring, or high-stakes decision-making. These risks increase when outputs are presented as objective measures of an individual's internal state. We caution against such uses and recommend that, when emotion predictions are used at all, systems should expose uncertainty (e.g., calibrated probabilities) and be accompanied by clear documentation of what the labels represent, how they were collected, and where disagreement is expected \citep[e.g.,][]{mohammad2022-ethics}.

Emotion categories, their lexical realizations, and their social meanings vary across languages and cultures. Models trained and evaluated on narrowly sampled annotator populations or single-language datasets may not transfer reliably to other contexts and may reproduce cultural biases. A practical implication of our work is that dataset documentation should report annotator composition, missingness patterns, and disagreement diagnostics, and that future annotation efforts should prioritize more diverse annotator pools and multilingual settings.

Because emotion annotation and interpretation are inherently tied to theories of affect, narrative, and communication, we encourage involving researchers beyond NLP (e.g., psychology, linguistics, HCI, digital humanities) in annotation design, guideline development, and evaluation choices. Disagreement-aware approaches should be informed by domain knowledge about when multiple interpretations are expected and meaningful, rather than treated as purely statistical noise.

Finally, any future release of materials derived from copyrighted literary sources will comply with licensing constraints. Where full text cannot be redistributed, we will document what can be shared (e.g., derived labels, code, and evaluation scripts) to support reproducibility while respecting rights holders.

\section*{Acknowledgments}
This paper was published in the Proceedings of the 1st Workshop on Computational Affective Science, CAS 2026, co-located with LREC 2026.

This work was supported by JSPS KAKENHI Grant Number 24K21058. \\ \\

\section{Bibliographical References}\label{sec:reference}
\bibliographystyle{lrec2026-natbib}
\bibliography{custom,custom-2,anthology-2}

\clearpage

\appendix
\onecolumn
\section{Inter-annotator emotion co-occurence}
\label{app:a}
\begin{figure}[h]
    \centering
    \includegraphics[width=0.7\linewidth]{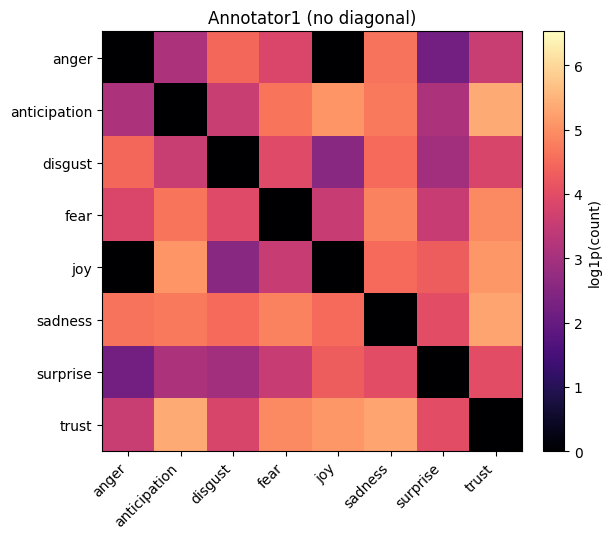}
    \caption{Annotator 1 emotion co-occurence}
    \label{fig:a1co}
\end{figure}

\begin{figure}[h]
    \centering
    \includegraphics[width=0.7\linewidth]{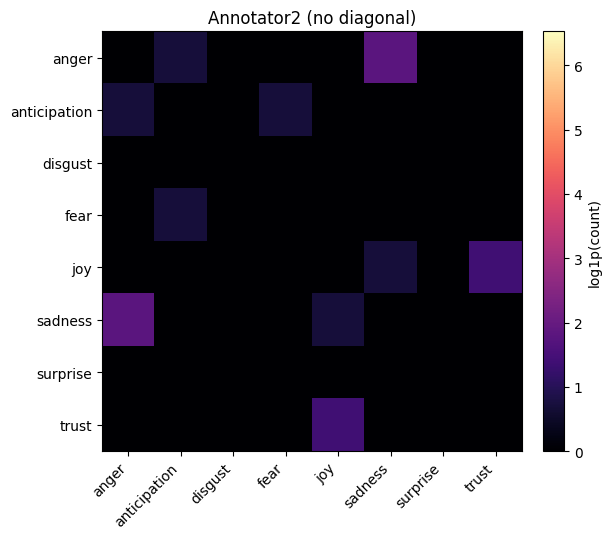}
    \caption{Annotator 2 emotion co-occurence}
    \label{fig:a2co}
\end{figure}

\begin{figure}[h]
    \centering
    \includegraphics[width=0.7\linewidth]{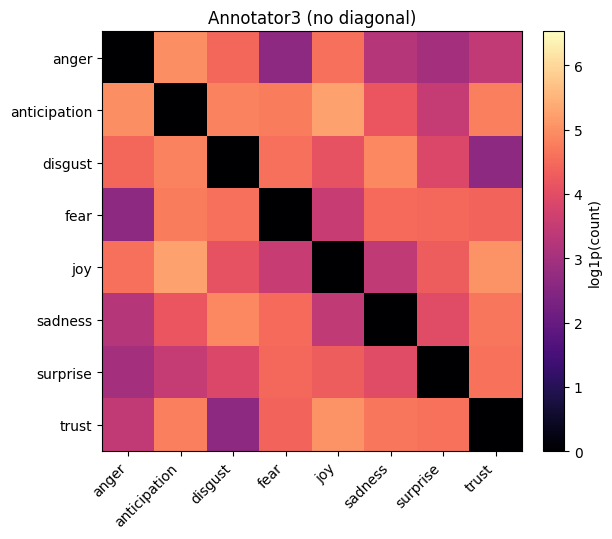}
    \caption{Annotator 3 emotion co-occurence}
    \label{fig:a3co}
\end{figure}
\clearpage
\section{Top other emotions}
\label{app:b}

\begin{table}[h]
\centering
\footnotesize
\setlength{\tabcolsep}{4pt}
\begin{tabular}{lr}
\toprule
\textbf{Label} & \textbf{Count} \\
\midrule
sentimentality & 6 \\
appreciation & 3 \\
unbelief & 3 \\
confusion & 3 \\
pessimism & 2 \\
disapproval & 2 \\
contempt & 2 \\
mild delight & 2 \\
hope & 2 \\
mild despair & 2 \\
mild contempt & 2 \\
shame & 2 \\
mild submission & 2 \\
determination & 2 \\
love & 1 \\
optimism & 1 \\
caring & 1 \\
protectiveness & 1 \\
childish exasperation or mild annoyance & 1 \\
remorse & 1 \\
acceptance & 1 \\
submission & 1 \\
mild disapproval & 1 \\
mild submission and love & 1 \\
mild remorse & 1 \\
loving pride & 1 \\
guilt & 1 \\
mild aggression plus optimism & 1 \\
suspicion & 1 \\
doubtfulness & 1 \\
\bottomrule
\end{tabular}
\caption{Out-of-taxonomy emotion labels provided in the free-text \textit{Other} field (frequency).}
\label{tab:other_label_freq}
\end{table}
Table~\ref{tab:other_label_freq} shows the top 30 most commonly occurring other labels outside of Plutchik's categories.
\begin{table}[ht]
\centering
\footnotesize
\setlength{\tabcolsep}{4pt}
\begin{tabular}{lr}
\toprule
\textbf{Emotion pairs} \\
\midrule
disapproval; contempt   \\
mild delight; hope   \\
sentimentality; pessimism   \\
love; optimism   \\
caring; protectiveness   \\
childish exasperation or mild annoyance   \\
unbelief; confusion   \\
shame, guilt   \\
sentimentality; mild submission   \\
mild aggression plus optimism   \\
\bottomrule
\end{tabular}
\caption{Top emotion co-occurrences for \textit{[Annotator X]} (off-diagonal counts).}
\label{tab:top_cooc_pairs_X}
\end{table}

Table~\ref{tab:top_cooc_pairs_X} shows the top co-occurring other category items
\end{document}